\documentclass{article} 
\usepackage{iclr2022_conference, times}

\usepackage{amsmath,amsfonts,bm}

\def\eqref#1{equation~\ref{#1}}

\def\1{\bm{1}}

\DeclareMathAlphabet{\mathsfit}{\encodingdefault}{\sfdefault}{m}{sl}
\SetMathAlphabet{\mathsfit}{bold}{\encodingdefault}{\sfdefault}{bx}{n}

\usepackage{hyperref}
\usepackage{url}
\usepackage{subfig}
\usepackage{tabularx}
\usepackage{graphicx}

\title{User-Level Membership Inference Attack against Metric Embedding Learning}

\author{Guoyao Li\\
  Zhejiang University\\
  Hangzhou, Zhejiang, China \\
  \texttt{guoyaoli@zju.edu.cn} \\
  \And
  Shahbaz Rezaei\\
  University of California\\
  Davis, CA, USA \\
  \texttt{srezaei@ucdavis.edu} \\
  \And
  Xin Liu\\
  University of California\\
  Davis, CA, USA \\
  \texttt{xinliu@ucdavis.edu} \\
}

\iclrfinalcopy 
\begin{document}

\maketitle

\begin{abstract}
Membership inference (MI) determines if a sample was part of a victim model training set. 
Recent development of MI attacks focus on record-level membership inference which limits their application in many real-world scenarios. For example, in the person re-identification task, the attacker (or investigator) is interested in determining if a user's images have been used during training or not. However, the exact training images might not be accessible to the attacker. In this paper, we develop a user-level MI attack where the goal is to find if any sample from the target user has been used during training even when no exact training sample is available to the attacker. We focus on metric embedding learning due to its dominance in person re-identification, where user-level MI attack is more sensible. We conduct an extensive evaluation on several datasets and show that our approach achieves high accuracy on user-level MI task.
\end{abstract}

\section{Introduction}

Membership inference (MI) attacks aim to identify whether a sample has been used during the training of a victim model or not. The existing research literature has primarily focused on record-level MI attack on classifiers and defense mechanisms against them.
Record-level MI attack has a major limitation: it assumes that the exact training samples are available at the inference time to conduct membership inference. For example, a privacy auditor may want to investigate if a user's images have been unlawfully used to train a model connected to a video surveillance camera by using MI attacks. The camera that records people's movements may constantly capture pictures and retrain a vision model. However, if a privacy auditor (using the technique of MI attacks) wants to identify the identity of people whose data is used to train the model (against their will), there is no practical way to retrieve those exact training images. To address this limitation, we focus on user-level membership inference, where the goal is to identify users whose images were used to train a model, given that the exact training images are not available.

Specifically, we investigate a scenario that differs from traditional record-level MI attacks in two key aspects: 1) We focus on a user-level MI attack where the goal is to identify if any image from a target person (user) has been used for training the victim model or not. The primary example of tasks for which the user-level MI attack is more sensible are person re-identification or face recognition. Here, we want to know if any image of a target person was a part of a training dataset, not just one specific image. 2) We focus on metric embedding learning rather than classifiers because they are widely used for person re-identification and face recognition. 

\begin{figure*}
\centering
\centering
\includegraphics[width=0.35\linewidth]{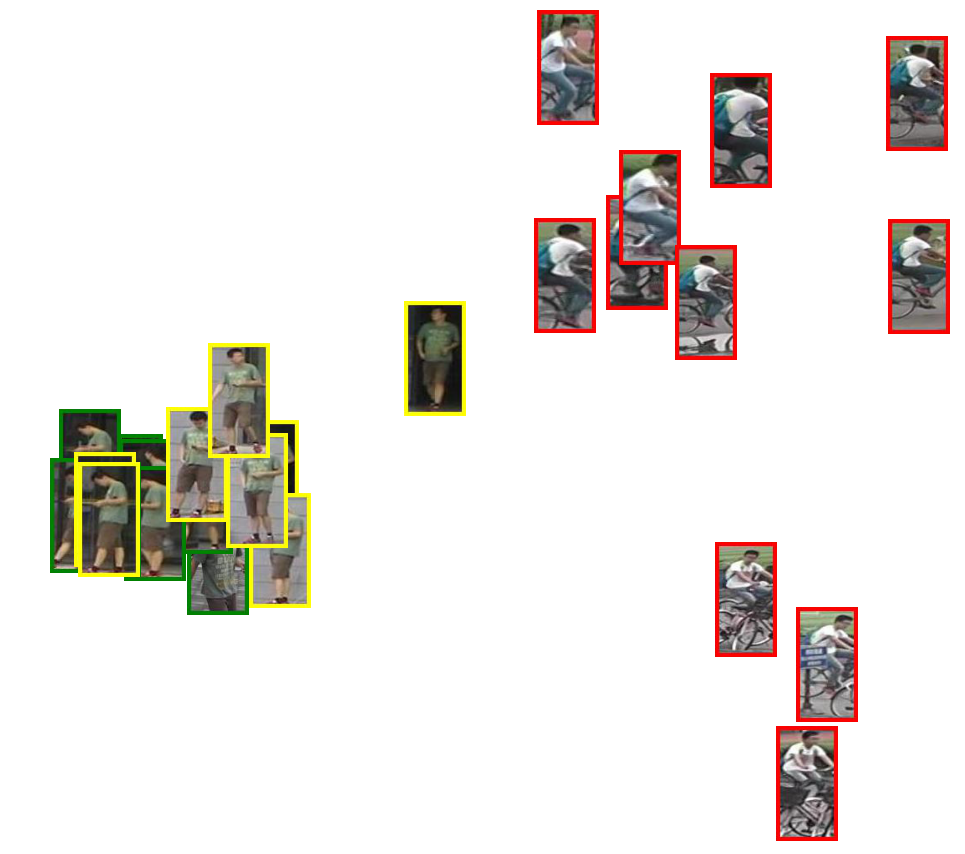}
\caption{Green: training members, yellow: non-training members, and red: non-member. The distances are computed based on the latent space embedding of a LuNet model.}
\label{fig-images}
\end{figure*}

These two differences result in two new challenges. 
First, in most existing work, the user-level setting is either undefined or ignored. For example, in CIFAR dataset, where the task is to classify objects or animals, the notion of a user or an entity beyond a record is not well-defined. Second, in metric embedding learning, the model output does not contain confidence values or labels based on which the majority of existing MI attacks are built. To address these two challenges, we propose a new user-level MI attack against metric embedding based on an \textbf{intuitive empirical observation}: users whose data has been used during training form more compact clusters in the latent space. As shown in Figure \ref{fig-images}, this observation holds both for training samples (green color) and other images of the same person that have not been used during training (yellow color), which solves the first challenge. Moreover, we focus on cluster properties in the latent space rather than on confidence output to address the second challenge.

In this paper, we introduce a user-level MI attack against metric embedding learning using properties of clusters in latent space. More specifically, we use average distance to the cluster's center and average pair-wise distance as features. We show that our attack achieves high accuracy even when the target model is probed with images of a training user that have not been used in the training, and therefore, we make the user-level MI attack viable.

\section{Background}
\subsection{Membership Inference}
The goal of membership inference is to identify whether a sample was part of a victim training model or not. Existing membership inference attacks, such as \cite{shokri2017membership, salem2019ml}, mainly focus on \textit{record-level MI attack} on classification tasks. The main intuition behind these MI attacks is that classification models are more confident on training samples than test samples, and hence the confidence values can be used to infer membership \citep{rezaei2020towards}.

In this paper, we focus on the \textit{user-level MI attack}, where the goal is to identify if any sample (images) from a target user has been used in the training. Here, the attacker might not have access to the exact training samples, but she can obtain other samples from the same user. This attack is more relevant in tasks where a user's identity is in danger of leaking, such as person re-identification. In the literature, there are only a few studies on user-level MI attacks. In \cite{miao2021audio}, the authors investigate MI attacks on speech recognition task to infer if any users' data (voice samples) have been used during training. In \cite{song2019auditing}, the authors propose a user-level MI attack on text generative models. None of the existing user-level MI attacks can be directly adopted for metric embedding learning scenario as discussed in detail in Sec.~\ref{subsec:setting}.

\subsection{Metric Embedding Learning}
The goal of metric embedding learning is to learn a mapping from a high-dimensional input space into a lower-dimensional latent space in which semantically similar inputs are closer \citep{hermans2017defense}. This includes variations of contrastive loss and triplet loss. In contrastive loss, two samples are taken as the input to a model, and the loss term aims to decrease (increase) the distance of the embeddings of these samples if they belong to similar (different) class(es). Here, samples from similar classes are called \textit{positive samples}, and samples from different classes are called \textit{negative samples}. The triplet loss takes three samples as input: an anchor, a positive sample w.r.t the anchor, and a negative sample w.r.t the anchor. It aims to push anchor and positive samples together while pulling the anchor and negative samples away. None of the existing MI attacks can be directly adopted for metric embedding learning because the outputs of metric embeddings are not confidence values. To the best of our knowledge, the only MI attack on metric embeddings is EncoderMI \citep{liu2021encodermi}. Simply put, it computes the closeness of a target image with its augmented versions in latent space as attack feature. However, it is a record-level MI attack, and we show that its extension to a user-level scenario leads to poor performance.

\section{Attack Method}
\subsection{Threat Model}

\textbf{Victim Model:}
In this paper, we mainly use the LuNet model with soft-margin batch hard loss \citep{hermans2017defense}, a variant of triplet loss, as a victim model due to its high accuracy and popularity. LuNet loss modifies the original triplet loss to efficiently choose the hardest positive and hardest negative samples for each anchor sample to improve the training. Note that our approach can be trivially extended to any other metric embedding learning because it uses the embedding as a black-box function.

\textbf{User-level Membership Inference:}
In contrast to record-level membership inference, where samples are categorized into members and non-members, in user-level membership inference we have three groups of samples: 1) \textit{training members} ($D^{t}_{m}$) are the samples from users that have been used during the training, 2) \textit{non-training members} ($D^{nt}_{m}$) are samples that have not been used during the training, but the identity of the corresponding users have been used via training member samples, and 3) \textit{non-members} ($D_{nm}$) are samples from users whose data has never been used during the training. Here, the goal is to identify non-training members as members without accessing training members, which is in general not available in record-level MI attacks.

\textbf{Attacker knowledge:}
We assume that the attacker has access to a set of non-training member samples and a set of non-members. However, the attacker does not know which sample belong to which set. The attacker does not necessarily need training members which is a more realistic assumption in comparison with record-level MI attacks where the exact training samples should be available to the attacker to identify members. Additionally, we assume that the attacker can query the black-box encoder to obtain the latent representation of samples.

\subsection{Feature Extraction}
\label{sec_features}
\textbf{Key intuition:} The key observation that allows an attacker to launch an MI attack against metric embeddings is that the images of the user whose data has been used during the training form a more compact cluster in the latent space of the victim model, as shown in Figure \ref{fig-images}. This includes both training members ($D^{t}_{m}$) and non-training members ($D^{nt}_{m}$).

\textbf{Attack features:} To use the key observation stated above, we need to measure the compactness of user's samples in latent space. To achieve this goal, we define two metrics: 1) average center-based distance ($C_u$), and 2) average pair-wise distance ($P_u$). Let's denote $E_v(.)$ as the victim model that outputs the latent representation. We use $x^i_u$ to denote the $i^{th}$ sample of a user, $u$. Given $m_u$ samples from the user $u$, average center-based distance is defined as follows:

\begin{equation}
    C_u = \frac{1} {m_u} \sum_{i=1}^{m_u}  d(x^i_u, \bar{x}_u),
\end{equation}

where $\bar{x}_u = \frac{1}{m_u} \sum_{i=1}^{m_u} x^i_u$, called the center of cluster, and d(.) is a distance measure. We use the L2 norm as the distance measure throughout this paper. Similarly, we define the average pair-wise distance as follows:

\begin{equation}
    P_u = \frac{1}{{m_u - 1}} \sum_{i=1}^{m_u-1}  \frac{\sum_{j=i+1}^{m_u} d(x^i_u, x^j_u) } {m_u-(i+1)},
\end{equation}

which obtains the average latent distance across all possible pairs of images of user $u$. Note that in contrast to existing record-level MI attacks, we cannot infer the membership of a user using only a single sample. To measure the compactness of a cluster, our attack requires multiple samples from the user.

\subsection{Attack Model Training}
Using the two attack features ($C_u, P_u$) described above as input to the attack model, we train an attack model to output the membership status of a target user. We adopt shadow model training strategy widely used in record-level MI attack proposed in \citep{shokri2017membership}. Simply put, we train multiple (shadow) models on the same task as the victim model, but with different data samples. Since the ground truth of members and non-members of the shadow models are known to the attacker, she can use the ground truth to train the attack model. The details of the shadow models and their dataset is explained in Section \ref{sec-evaluation}.

\section{Evaluation}
\label{sec-evaluation}
\subsection{Experimental Settings}
\label{subsec:setting}

\textbf{Dataset}: We use Market-1501 \citep{zheng2015scalable} and PRID-2011 \citep{hirzer11}. Market-1501 is a benchmark frequently used to evaluate person re-identification models. After excluding duplicates, distractors and junks, we have 26051 labeled images of 1501 users. PRID-2011 consists of images extracted from multiple person trajectories. After excluding duplicates, we have 71657 labeled images of 934 users.

\textbf{Victim model}: We choose LuNet with soft-margin batch hard loss by \cite{hermans2017defense} as our victim model, which is trained on $D^{t}_{m}$. For Market-1501 and PRID-2011, we randomly select $D^{t}_{m}$, $D^{nt}_{m}$, and $D_{nm}$ from the dataset. $D^{t}_{m}$ and $D^{nt}_{m}$ includes non-overlapping images from the same 150 memebrs. $D_{nm}$ includes images of 150 non-members, who do not overlap with the members.
The remaining images are used as the shadow dataset, $D_s$.

\textbf{Shadow models}: For each shadow model, we randomly select shadow training members, shadow non-training members, and shadow non-members from the shadow dataset, $D_s$. We train shadow models on shadow training member set. Here, shadow model architecture is the same as the victim model architecture, both in our attack and \cite{liu2021encodermi} with which we compare our attack. We train 10 and 100 shadow models for PRID-2011 and Market-1501 datasets, respectively.

\textbf{Attack model}: Our attack model is a shallow neural network with 3 fully connected layers. The input features are the average center-based distance ($C_u$) and average pair-wise distance ($P_u$) as described in Section \ref{sec_features}. Throughout our evaluation, we always use the same number of images to obtain these two features. We train the attack model with the shadow dataset. We repeat each experiment 5 times and report the average and standard deviation.

\textbf{Baselines}: To the best of our knowledge, there is no user-level MI attack on metric embedding learning. The two user-level MI attacks in literature \citep{song2019auditing, miao2021audio} require generative models where the victim model's output is a word. Hence, there is no trivial way to adopt them for metric embedding scenario. Moreover, the majority of record-level MI attacks on classifiers rely on confidence values which is not available when using metric embedding. Hence, there is no trivial way to adopt them here. However, we can adopt record-level MI attacks on metric embedding to the user-level scenario with a minor adjustment. There is only one attack that satisfy this condition, called EncoderMI \citep{liu2021encodermi}. To adopt for the user-level MI scenario, we launch their record-level MI attack on all samples of a user and then we perform majority voting.


\begin{table}[t]
\caption{Performance comparison of user-level MI attacks on metric embeddings. 
}
\label{tbl-base}
\begin{center}
\resizebox{\textwidth}{!}{
\begin{tabular}{lllllll}
\multicolumn{1}{c}{\bf MIA method}  &\multicolumn{2}{c}{\bf Accuracy} &\multicolumn{2}{c}{\bf Precision} &\multicolumn{2}{c}{\bf Recall}
\\
\multicolumn{1}{c}{}  &\multicolumn{1}{c}{\bf Market} &\multicolumn{1}{c}{\bf PRID} &\multicolumn{1}{c}{\bf Market} &\multicolumn{1}{c}{\bf PRID} &\multicolumn{1}{c}{\bf Market} &\multicolumn{1}{c}{\bf PRID} 
\\ \hline \\
Our user-level MIA  &\textbf{66.87} $\pm$ 1.87  &\textbf{74.27} $\pm$ 0.83  &75.25 $\pm$ 0.54  &69.80 $\pm$ 1.35  &50.27 $\pm$ 5.51  &85.73 $\pm$ 2.94   \\
EncoderMI (unknown augmentations)     &52.00 $\pm$ 1.56  &52.67 $\pm$ 2.14  &54.28  $\pm$ 3.32   &51.06 $\pm$ 3.02   &46.67 $\pm$ 30.94  &63.33 $\pm$ 32.49      \\
EncoderMI (full knowledge)     &66.00 $\pm$ 1.21   &69.60 $\pm$ 3.12  &63.62 $\pm$ 3.55  &65.20 $\pm$ 3.96   &77.60 $\pm$ 10.55  &86.27 $\pm$  6.02\\
\end{tabular}}
\end{center}
\end{table}
\subsection{Performance Comparison}

Table \ref{tbl-base} shows the performance comparison between our attack and EncoderMI. Here, we only use non-training members and non-members for the evaluation purpose. EncoderMI computes the closeness of the target sample with its augmented variants as features. When the exact data augmentations used by the victim model are not known to the attacker, it chooses a fixed set of augmentations following the original setting of EncoderMI paper. In this case, the EncoderMI performs close to random guess (the second row). However, when all data augmentations during victim model training are known to the attacker, EncoderMI performs better (the third row). Despite such an unrealistic advantage to the EncoderMI, it still cannot outperform our approach.

\subsection{Access to some training images}
In the previous section, we assumed that only the non-training member samples are available to the user-level MIA. In cases where some training member samples are available, we expect to achieve even better performance. As shown in Table \ref{tbl-training-data}, by increasing the number of training members available to the attacker, we can significantly improve the user-level MI accuracy. 


\begin{table}[t]
\caption{User-level MIA performance when some portion of the training samples are available to the attacker.}
\label{tbl-training-data}
\begin{center}
\resizebox{\textwidth}{!}{
\begin{tabular}{lllllll}
\multicolumn{1}{c}{\bf Proportion of training}  &\multicolumn{2}{c}{\bf Accuracy} &\multicolumn{2}{c}{\bf Precision} &\multicolumn{2}{c}{\bf Recall}
\\
\multicolumn{1}{c}{}  &\multicolumn{1}{c}{\bf Market} &\multicolumn{1}{c}{\bf PRID} &\multicolumn{1}{c}{\bf Market} &\multicolumn{1}{c}{\bf PRID} &\multicolumn{1}{c}{\bf Market} &\multicolumn{1}{c}{\bf PRID} 
\\ \hline \\
0\%     &66.87 $\pm$ 1.87    &74.27 $\pm$ 0.83    &75.25 $\pm$ 0.54
        &69.80 $\pm$ 1.35    &50.27 $\pm$ 5.51    &85.73 $\pm$ 2.94 \\
25\%    &74.60 $\pm$ 0.25    &76.33 $\pm$ 0.30    &79.96 $\pm$ 1.18  
        &70.78 $\pm$ 1.20    &65.73 $\pm$ 2.33    &89.87 $\pm$ 3.08 \\
50\%    &81.87 $\pm$ 0.69    &78.53 $\pm$ 0.54    &82.97 $\pm$ 1.05  
        &71.76 $\pm$ 1.36    &80.27 $\pm$ 3.00    &94.27 $\pm$ 2.25 \\
75\%    &90.00 $\pm$ 0.52    &78.40 $\pm$ 0.65    &85.41 $\pm$ 1.26  
        &71.70 $\pm$ 1.40    &96.53 $\pm$ 0.98    &94.00 $\pm$ 2.11 \\
100\%   &91.73 $\pm$ 0.93    &78.07 $\pm$ 1.00    &85.83 $\pm$ 1.35  
        &71.56 $\pm$ 1.52    &100.0 $\pm$ 0.00    &93.33 $\pm$ 1.89 \\
\end{tabular}}
\end{center}
\end{table}

\subsection{Ablation Analysis}
Table \ref{tbl-ablation} illustrates the effect of each attack feature on user-level MIA. Although the highest accuracy is achieved when both features are used, the difference is not significant. Hence, the attacker can also use a single feature to reduce the computation overhead.


\begin{table}[t]
\caption{User-level MIA performance evaluation using different set of features.}
\label{tbl-ablation}
\begin{center}
\resizebox{\textwidth}{!}{
\begin{tabular}{lllllll}
\multicolumn{1}{c}{\bf Input Features}  &\multicolumn{2}{c}{\bf Accuracy} &\multicolumn{2}{c}{\bf Precision} &\multicolumn{2}{c}{\bf Recall}
\\
\multicolumn{1}{c}{}  &\multicolumn{1}{c}{\bf Market} &\multicolumn{1}{c}{\bf PRID} &\multicolumn{1}{c}{\bf Market} &\multicolumn{1}{c}{\bf PRID} &\multicolumn{1}{c}{\bf Market} &\multicolumn{1}{c}{\bf PRID} 
\\ \hline \\
$(C_u)$         &65.80 $\pm$ 3.39    &73.13 $\pm$ 0.45    &75.67 $\pm$ 1.29  
                &68.17 $\pm$ 0.27    &46.93 $\pm$ 11.23   &86.80 $\pm$ 2.12\\
$(P_u)$         &66.67 $\pm$ 2.32    &73.53 $\pm$ 0.83    &74.40 $\pm$ 1.23  
                &69.20 $\pm$ 1.53    &51.07 $\pm$ 8.42    &85.07 $\pm$ 3.34\\
$(C_u$,$P_u)$   &66.87 $\pm$ 1.87    &74.27 $\pm$ 0.83    &75.25 $\pm$ 0.54
                &69.80 $\pm$ 1.35    &50.27 $\pm$ 5.51    &85.73 $\pm$ 2.94 \\
\end{tabular}}
\end{center}
\end{table}

\section{Conclusion}
In this paper, we propose a user-level MI attack on metric embedding learning. Our attack differs from most existing MI attacks in two aspects: First, we focus on the user-level MI attack which is more practical in tasks where the exact training data samples used in training are not available. Second, we focus on metric embedding learning scenario where the existing confidence-based MI attacks do not work. In contrast with existing MI attacks, we use a measure of compactness of clusters in embedding space to identify membership, and consequently, obviate the need to access confidence values. Our attack achieves the state-of-the-art performance in several datasets, where user-level MI attack is of paramount importance.


\bibliography{iclr2022_conference}

\begin{thebibliography}{9}
\providecommand{\natexlab}[1]{#1}
\providecommand{\url}[1]{\texttt{#1}}
\expandafter\ifx\csname urlstyle\endcsname\relax
  \providecommand{\doi}[1]{doi: #1}\else
  \providecommand{\doi}{doi: \begingroup \urlstyle{rm}\Url}\fi

\bibitem[Hermans et~al.(2017)Hermans, Beyer, and Leibe]{hermans2017defense}
Alexander Hermans, Lucas Beyer, and Bastian Leibe.
\newblock In defense of the triplet loss for person re-identification.
\newblock \emph{arXiv preprint arXiv:1703.07737}, 2017.

\bibitem[Hirzer et~al.(2011)Hirzer, Beleznai, Roth, and Bischof]{hirzer11}
Martin Hirzer, Csaba Beleznai, Peter~M. Roth, and Horst Bischof.
\newblock {Person Re-Identification by Descriptive and Discriminative
  Classification}.
\newblock In \emph{{Proc. Scandinavian Conference on Image Analysis (SCIA)}},
  2011.

\bibitem[Liu et~al.(2021)Liu, Jia, Qu, and Gong]{liu2021encodermi}
Hongbin Liu, Jinyuan Jia, Wenjie Qu, and Neil~Zhenqiang Gong.
\newblock Encodermi: Membership inference against pre-trained encoders in
  contrastive learning.
\newblock In \emph{Proceedings of the 2021 ACM SIGSAC Conference on Computer
  and Communications Security}, pp.\  2081--2095, 2021.

\bibitem[Miao et~al.(2021)Miao, Minhui, Chen, Pan, Zhang, Zhao, Kaafar, and
  Xiang]{miao2021audio}
Yuantian Miao, Xue Minhui, Chao Chen, Lei Pan, Jun Zhang, Benjamin Zi~Hao Zhao,
  Dali Kaafar, and Yang Xiang.
\newblock The audio auditor: user-level membership inference in internet of
  things voice services.
\newblock \emph{Proceedings on Privacy Enhancing Technologies}, 2021:\penalty0
  209--228, 2021.

\bibitem[Rezaei \& Liu(2021)Rezaei and Liu]{rezaei2020towards}
Shahbaz Rezaei and Xin Liu.
\newblock On the difficulty of membership inference attacks.
\newblock In \emph{Proceedings of the IEEE/CVF Conference on Computer Vision
  and Pattern Recognition}, 2021.

\bibitem[Salem et~al.(2019)Salem, Zhang, Humbert, Fritz, and
  Backes]{salem2019ml}
Ahmed Salem, Yang Zhang, Mathias Humbert, Mario Fritz, and Michael Backes.
\newblock Ml-leaks: Model and data independent membership inference attacks and
  defenses on machine learning models.
\newblock In \emph{Network and Distributed Systems Security Symposium 2019}.
  Internet Society, 2019.

\bibitem[Shokri et~al.(2017)Shokri, Stronati, Song, and
  Shmatikov]{shokri2017membership}
Reza Shokri, Marco Stronati, Congzheng Song, and Vitaly Shmatikov.
\newblock Membership inference attacks against machine learning models.
\newblock In \emph{2017 IEEE symposium on security and privacy (SP)}, pp.\
  3--18. IEEE, 2017.

\bibitem[Song \& Shmatikov(2019)Song and Shmatikov]{song2019auditing}
Congzheng Song and Vitaly Shmatikov.
\newblock Auditing data provenance in text-generation models.
\newblock In \emph{Proceedings of the 25th ACM SIGKDD International Conference
  on Knowledge Discovery \& Data Mining}, pp.\  196--206, 2019.

\bibitem[Zheng et~al.(2015)Zheng, Shen, Tian, Wang, Wang, and
  Tian]{zheng2015scalable}
Liang Zheng, Liyue Shen, Lu~Tian, Shengjin Wang, Jingdong Wang, and Qi~Tian.
\newblock Scalable person re-identification: A benchmark.
\newblock In \emph{Computer Vision, IEEE International Conference on}, 2015.

\end{thebibliography}
\bibliographystyle{iclr2022_conference}

\appendix
\section{Appendix}

\subsection{Effect of number of training samples versus MI attack}
Intuitively, as the number of training samples for a user increases, we expect the metric embedding process to push those images more towards each other. In other words, as the number of training samples for a user increases, it presents a more compact cluster in the latent space. Table \ref{tbl-training-size} shows our user-level MIA recall on different group of users with different number of training samples. Clearly, our MI attack is more successful on users with larger number of training samples. This is somehow in contrast with record-level MIA on classifiers where more training data is often construed as less memorization and, hence, less privacy leakage.



\begin{table}[hbt!]
\caption{User-level MIA's recall on groups with different number of training images per person.}
\label{tbl-training-size}
\begin{center}
\begin{tabular}{lllll}
\multicolumn{1}{c}{\bf Group}  
&\multicolumn{2}{c}{\bf Market} &\multicolumn{2}{c}{\bf PRID}
\\
\multicolumn{1}{c}{}  &\multicolumn{1}{c}{\bf Number of Images} &\multicolumn{1}{c}{\bf Recall}
&\multicolumn{1}{c}{\bf Number of Images} &\multicolumn{1}{c}{\bf Recall}
\\ \hline \\
1   &$22 \leq n \leq 63$        &69.33 $\pm$ 5.73  &$123 \leq n \leq 445$   &96.77 $\pm$ 2.04   \\
2   &$17 \leq n \leq 21$        &60.00 $\pm$ 7.43  &$102 \leq n \leq 112$   &85.81 $\pm$ 5.62   \\
3   &$14 \leq n \leq 16$        &40.83 $\pm$ 4.86  &$88 \leq n \leq 101$    &84.14 $\pm$ 1.69   \\
4   &$11 \leq n \leq 13$        &36.47 $\pm$ 5.76  &$78 \leq n \leq 87$     &85.33 $\pm$ 1.63   \\
5   &$8 \leq n \leq 10$         &45.45 $\pm$ 5.07  &$66 \leq n \leq 77$     &75.86 $\pm$ 4.88   \\
\end{tabular}
\end{center}
\end{table}

\end{document}